\documentclass[10pt,twocolumn,letterpaper]{article}

\usepackage[pagenumbers]{cvpr} 

\usepackage[dvipsnames]{xcolor}

\usepackage{amssymb}
\usepackage{pifont}
\definecolor{cvprblue}{rgb}{0.21,0.49,0.74}
\usepackage[pagebackref,breaklinks,colorlinks,citecolor=cvprblue]{hyperref}
\usepackage{xcolor}
\usepackage{colortbl}
\newcommand{\rowA}{\rowcolor{gray!15}}
\usepackage{multirow}
\usepackage{array}
\newcolumntype{M}[1]{>{\centering\arraybackslash}p{#1}}
\definecolor{myblue}{RGB}{1,137,157}
\definecolor{mygreen}{RGB}{177,196,77}
\definecolor{mypink}{RGB}{238,176,175}
\usepackage{caption}
\usepackage{float}
\usepackage[symbol*]{footmisc}

\def\paperID{11919} 
\def\confName{CVPR}
\def\confYear{2024}

\title{\shortname: \longname}

\author{%
  Chandradeep Pokhariya\textsuperscript{1} \textsuperscript{*} \and
  Ishaan Nikhil Shah\textsuperscript{1} \textsuperscript{**} \and
  Angela Xing\textsuperscript{2} \textsuperscript{**} \and
  Zekun Li\textsuperscript{2} \and
  Kefan Chen\textsuperscript{2} \and 
  Avinash Sharma\textsuperscript{1} \and Srinath Sridhar\textsuperscript{2} \and
\centerline{\textsuperscript{1}IIIT Hyderabad \quad \textsuperscript{2}Brown University
} \\
\centerline{
\href{https://ivl.cs.brown.edu/research/manus.html}{ivl.cs.brown.edu/research/manus}
}
}

\usepackage{xspace}
\newcommand{\shortname}{MANUS\xspace}
\newcommand{\longname}{Markerless Grasp Capture using Articulated 3D Gaussians\xspace}
\newcommand{\longnameu}{\textbf{\underline{Ma}}rkerless Ha\textbf{\underline{n}}d-Object Grasp Capture \textbf{\underline{u}}sing Articulated 3D Gau\textbf{\underline{s}}sians\xspace}
\newcommand{\handmodel}{MANUS-Hand\xspace}
\newcommand{\DatasetName}{MANUS-Grasps\xspace}
\newcommand{\parahead}[1]{\noindent\textbf{#1}:\ }

\newenvironment{packed_itemize}
{\begin{itemize}
    \setlength{\itemsep}{1pt}
    \setlength{\parskip}{0pt}
    \setlength{\parsep}{0pt}
}{\end{itemize}}

\newcommand{\filluptopage}[1]{%
  \clearpage
  \loop\ifnum\value{page}<#1\relax
    \null\clearpage
  \repeat
  \loop\ifnum\value{page}=#1\relax
    \null\clearpage
  \repeat
}

\begin{document}

\newcommand{\teaserCaption}{ }

\twocolumn[{
    \renewcommand\twocolumn[1][]{#1}
    \maketitle
    
    \vspace{-0.5 cm}
    \centering
    \begin{minipage}{1.00\textwidth}
        \centering 
        \includegraphics[width=\linewidth]{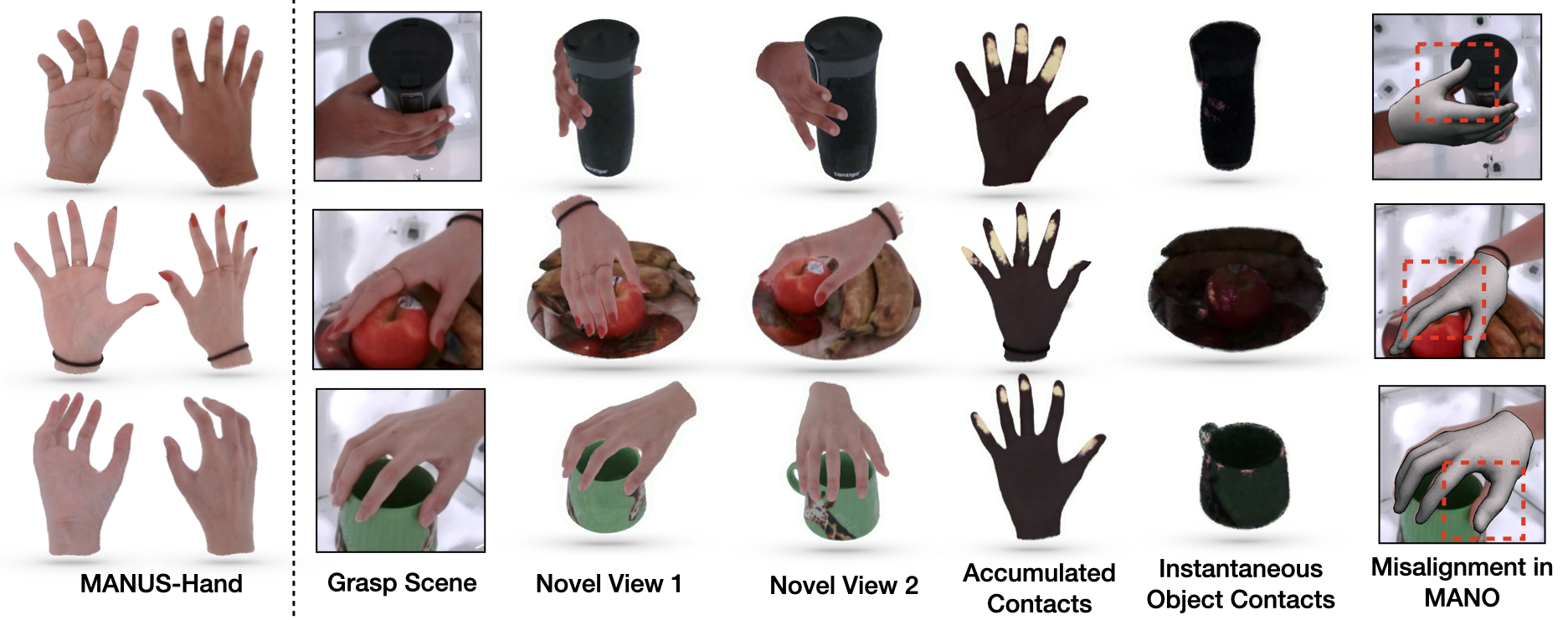}
    \end{minipage}
    \vspace{-0.5 em}
    \captionsetup{type=figure}
    \vspace{-2mm}
    \captionof{figure}{\teaserCaption We introduce \textbf{\shortname}, a novel markerless approach for capturing grasps by employing an articulated 3D Gaussian representation to accurately model hand shapes. This approach improves contact estimation accuracy in comparison to other template-based approaches when evaluated against ground truth contacts.}
    \label{fig:teaser}
    \vspace{4mm}
}]
\footnotetext{** Equal Contribution}
\footnotetext{* Work was done while at Brown University}
\begin{abstract}
Understanding how we grasp objects with our hands has important applications in areas like robotics and mixed reality.
However, this challenging problem requires accurate modeling of the contact between hands and objects.
To capture grasps, existing methods use skeletons, meshes, or parametric models that does not represent hand shape accurately resulting in inaccurate contacts.
We present \shortname, a method for \lowercase{\longnameu}.
We build a novel articulated 3D Gaussians representation that extends 3D Gaussian splatting~\cite{Kerbl20233DGS} for high-fidelity representation of articulating hands.
Since our representation uses Gaussian primitives optimized from the multi-view pixel-aligned losses, it enables us to efficiently and accurately estimate contacts between the hand and the object.
For the most accurate results, our method requires tens of camera views that current datasets do not provide.
We therefore build \DatasetName, a new dataset that contains hand-object grasps viewed from 50+ cameras across 30+ scenes, 3 subjects, and comprising over 7M frames.
In addition to extensive qualitative results, we also show that our method outperforms others on a quantitative contact evaluation method that uses paint transfer from the object to the hand.
\vspace{-0.2in}
\end{abstract}

\section{Introduction}
\label{sec:intro}

Every day, the average person effortlessly grasps more than a hundred different objects~\cite{zuccotti2015every,zheng2011investigation}.
This seemingly routine act of grasping poses a significant
challenge for machines, as is evident from the extensive research on this topic in computer vision~\cite{erol2007vision} and robotics~\cite{bicchi2000robotic,bohg2013data}.
High-fidelity capture of natural human grasps could unlock new applications in areas like robotics and mixed reality, but 
this challenging problem first requires us to accurately \textbf{estimate the contact} between the hand and the object~\cite{brahmbhatt2019contactdb}.

Previous work has addressed this problem by using gloves or special sensors~\cite{garcia2018first,pham2017hand}, but these devices are cumbersome and restrict hand movement.
Therefore, a large body of work has focused on \textbf{markerless grasp capture} using one or more cameras~\cite{hampali2020honnotate,sridhar2016real,ballan2012motion,chao2021dexycb,brahmbhatt2020contactpose}.

Most of these methods use skeletons~\cite{hampali2020honnotate}, meshes~\cite{ballan2012motion}, or parametric models~\cite{MANO:SIGGRAPHASIA:2017,joo2018total} to model the hand and object.
Although these representations are flexible and easy to use, they often cannot accurately model hand shape
resulting in reduced contact accuracy (see \Cref{fig:teaser}).
Recently, articulated neural implicit representations~\cite{mildenhall2020nerf,corona2022lisa,mundra2023livehand} have been proposed as alternatives, but modeling contact in implicit representations is challenging and requires expensive sampling.

To overcome these limitations, we introduce \textbf{\shortname}, a method for \longnameu.
The key component of \shortname is a 3D Gaussian splatting~\cite{Kerbl20233DGS} approach to build \textbf{\handmodel}, an articulated hand model composed of 3D Gaussians that make it faster to optimize and infer than many implicitly-represented models.
Similarly, we also capture the object using static 3D Gaussians.
Since both \handmodel and the object are modeled using Gaussians primitives with explicit positions and orientations, we can efficiently compute both \emph{instantaneous} and \emph{accumulated} contacts between them (see \Cref{sec:grasp_capture}).
When trained on datasets with tens of camera views, our method can accurately capture grasps since 3D Gaussians promote accurate pixel-level alignment resulting in more precise shape and contact estimation compared to existing methods.

Previous datasets~\cite{zimmermann2019freihand, hasson:hal-02429093, hampali2020honnotate,liu2022hoi4d,brahmbhatt2019contactdb,GRAB:2020,fan2023arctic,keypoint_transformer} have been instrumental in addressing the grasp capture problem but
(1)~they use specialized hardware (heat-sensitive cameras~\cite{brahmbhatt2019contactdb}, or markers~\cite{GRAB:2020}) to capture hand-object grasps, making it hard to scale,
(2)~RGB camera-only datasets~\cite{fan2023arctic, kwon2021h2o, brahmbhatt2020contactpose, chao2021dexycb}, 
contain only a few views with occlusions making it hard to learn accurate contacts, and
(3)~they rely on the parametric models or skeletons to estimate contacts resulting in inaccurate contacts.
\textbf{Our main insight is that accurate contact modeling is much easier with a large number of camera views that reduce the effect of (self-)occlusions.}
Therefore, we curated a one-of-a-kind real-world multi-view RGB dataset, \textbf{\DatasetName}, comprising over \textbf{\~7M frames} captured using 50+ high-framerate cameras, providing a full 360-degree coverage of grasp sequences occurring in over 30 diverse everyday scenarios.
In addition, this dataset contains 15 evaluation sequences that employ wet paint on objects to leave a contact residue on the hand~\cite{kamakura1980patterns} providing a natural way to evaluate contact quality without additional equipment or annotation.
We show extensive experiments ablating and justifying different components of \handmodel, as well as the \shortname grasping method.
In addition, we also provide a new metric of contact quality to assess the performance of \shortname against template-based methods.
While our method is not designed for photorealism, we observe that the captured grasping sequences are comparable in visual quality to the best implicit hand models.

To summarize, our contributions include: 
\begin{packed_itemize}
    \item \textbf{\handmodel}, a new efficient representation for articulated hands that uses 3D Gaussian splatting for accurate shape and appearance representation.
    \item \textbf{\shortname}, a method that uses \handmodel and a 3D Gaussian representation of the object to accurately model contacts.
    \item \textbf{\DatasetName}, a large real-world multi-view RGB grasp dataset with over \~7M frames from 50+ cameras, providing full 360-degree coverage of grasps in over 30 diverse everyday life scenarios.
    \item A unique and novel approach to validate contact accuracy using \textbf{paint transfer} between the object and the hand.
\end{packed_itemize}

\section{Related Work}
\label{sec:literature}
%
%
\parahead{Representations}
Skeletons and collections of shape primitives were some of the first representations to be used for hand--object interaction modeling~\cite{pham2017hand,sridhar2016real}, but these representations are often not accurate enough for contact estimation.
Meshes~\cite{ballan2012motion} and parametric models~\cite{MANO:SIGGRAPHASIA:2017,joo2018total} are currently the most popular alternatives but can also be misaligned with observations due to their lower-dimensional representation (see \Cref{fig:teaser}).

Coordinate-based implicit neural networks, or neural fields~\cite{xie2022neuralfields}, have shown great promise in accurately modeling shape and appearance in static scenes~\cite{mildenhall2020nerf, Lombardi:2019, NEURIPS2019_b5dc4e5d,Occupancy, Park_2019_CVPR, chen2018implicit_decoder,yariv2021volume,wang2021neus,mueller2022instant,yu2021plenoxels,chen2022tensorf,Kerbl20233DGS} as well as dynamic scenes~\cite{dynerf,yoon2020novel,yan2023nerf,wang2023mixed, kplanes_2023,luiten2023dynamic}.
Several methods specifically address articulated shapes~\cite{li2022tava} like human bodies~\cite{liu2021neural, peng2021neural, peng2021animatable, weng_humannerf_2022_cvpr, li2022tava}, or hands~\cite{corona2022lisa,li2022nimble,mundra2023livehand,qian2020html,Karunratanakul2022HARPPH}.
However, they use representations that are inefficient for sampling and contact estimation.
In contrast, we propose a new articulated neural field representation that extends 3D Gaussian splatting~\cite{Kerbl20233DGS} to hands enabling efficient training/inference and contact estimation.

\parahead{Hand-Object Interaction Capture}
Previous work has attempted to model hand-object interactions using skeletons~\cite{hampali2020honnotate,kwon2021h2o}, or customized meshes~\cite{ballan2012motion} as the hand representation without explicitly estimating contacts.
Most other work~\cite{chao2021dexycb,GRAB:2020,fan2023arctic,hasson:hal-02429093,liu2022hoi4d} uses MANO in combination with mocap, or one or more camera views.
While it becomes easier to estimate contact with a parametric mesh model, misalignments are still common (see \Cref{fig:teaser}).
To overcome the difficulty of accurate contact estimation, some methods resort to physical simulation~\cite{christen2022dgrasp,graspd,zhang2023artigrasp}, but these are limited to synthetic grasps only.
In contrast, we propose a template-free articulated 3D Gaussian splatting model that provides a natural way to estimate accurate contacts.

\parahead{Grasp Datasets}
Datasets for human grasps are challenging to obtain because they need specialized hardware, extensive annotation, and significant post-processing to make them useful.
Some datasets use markers or special gloves to track the hand and object~\cite{bernardin2005sensor,garcia2018first,delpretoactionnet,taheri2020grab}
but this hinders natural hand motion and introduces changes in image appearance.
Synthetic datasets~\cite{OccludedHands_ICCV2017,GANeratedHands_CVPR2018,hasson:hal-02429093} suffer from a domain gap that makes it challenging to generalize to real data.
Therefore, work has focused on manual annotations~\cite{sridhar2016real,ballan2012motion,bullock2015yale,rogez2015understanding}, optimization~\cite{hampali2020honnotate}, or automatic annotation~\cite{simon2017hand,chao2021dexycb} from RGB or depth.
Many of these datasets provide only 3D hand poses and lack information about contacts.
Other datasets like InterHand2.6M~\cite{moon2020interhand2,zimmermann2019freihand} are limited to hands only without any objects, while others~\cite{shan2020understanding} focus on 2D understanding only.
Addressing these limitations, HOnnotate~\cite{hampali2020honnotate} introduces a markerless system for automatically annotating frames across 77K frames. 
However, the variety of objects and grasps in this dataset is somewhat limited.
ContactDB~\cite{brahmbhatt2019contactdb} and ContactPose~\cite{brahmbhatt2020contactpose} address this limitation targets a broader variety of grasps.
While ContactDB is captured using thermal imaging, ContactPose uses multi-view RGB-D data.
Nonetheless, both methods are restricted to 3D hand poses, use non-realistic objects, and lack sufficient views for neural fields.

In contrast, we introduce \DatasetName that includes diverse grasps from 50+ cameras capturing at 120~FPS specifically to support neural field methods.
In total, we provide over \~7M frames with ground truth camera poses, segmentation, and estimated contacts.

\begin{figure*}[h!]
  \centering
  \includegraphics[width=\linewidth]{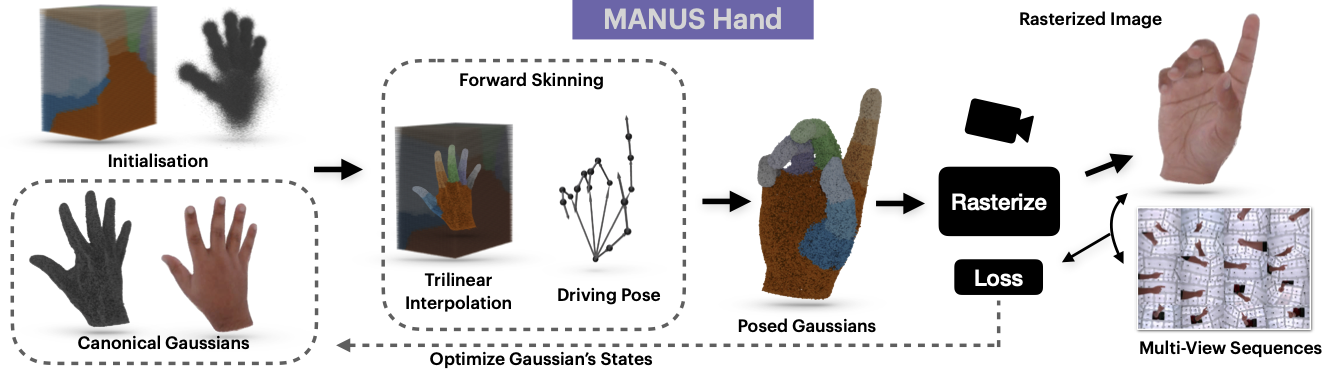}
  \caption{\textbf{\handmodel} is a template-free, articulable hand model learned from multi-view hand sequences which utilizes 3D Gaussian splatting representation for accurate modelling of the shape and appearance of hands.}
  
  \label{fig:arg_hand}
\end{figure*}

\begin{figure}[h!]
  \centering \includegraphics[width=\linewidth]{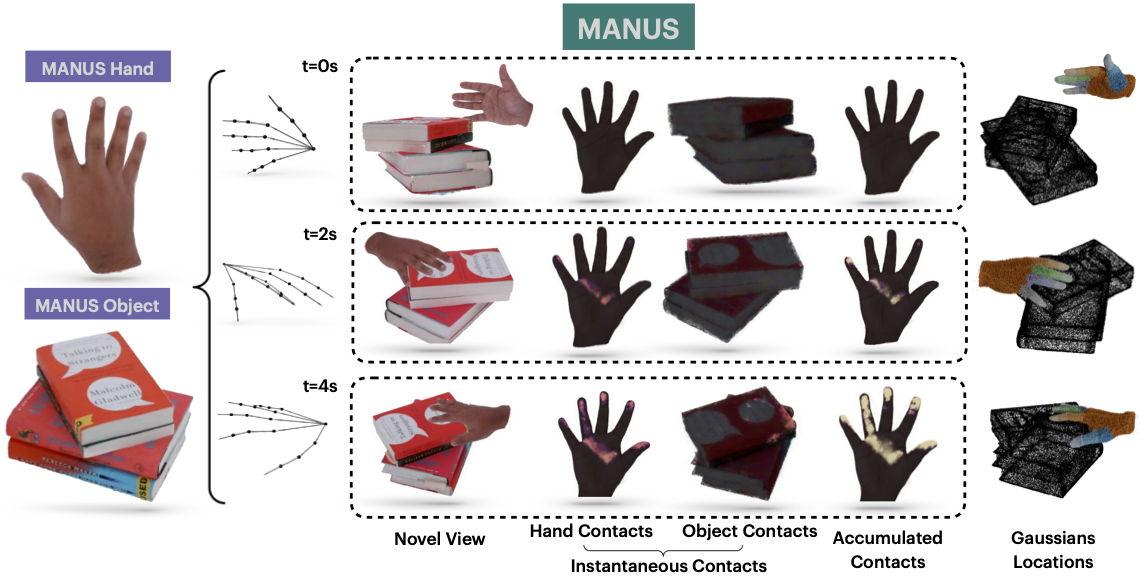}
  \caption{\textbf{\shortname} leverages a driving pose to get \handmodel in grasp scene. It is combined with an object model to get instantaneous and accumulated contacts between the two.
  }
  \label{fig:arg_mags}
  \vspace{-0.2in}
\end{figure}

{
\begin{table}[t!]
  \centering
  \begin{tabular}{M{2.5cm}@{}M{2.5cm}@{}M{2cm}}
    \toprule
    {Dataset} & {\#N Images (Views)} & {Annot. Type}\\
    \midrule
    \rowA \multicolumn{3}{l}{\textcolor{darkgray}{w/o Contacts Annotation}}   \\
    \textcolor{myblue}{H2O-3D} \cite{keypoint_transformer}           & 76k    (5)     & multi-kinect  \\
    \textcolor{myblue}{FHPA} \cite{garcia2018first}                  & 105k    (1)     & magnetic     \\
    \textcolor{myblue}{HOI4D}  \cite{liu2022hoi4d}                   & 2.4M    (1)     & single-manual \\
    \textcolor{mygreen}{FreiHand} \cite{zimmermann2019freihand}       & 37k     (8)     & semi-auto     \\
    \textcolor{mygreen}{HO3D} \cite{hampali2020honnotate}             & 78k     (1-5)   & multi-kinect    \\
    \textcolor{mygreen}{DexYCB} \cite{chao2021dexycb}                 & 582k    (8)     & multi-manual       \\
    \textcolor{mygreen}{ARCTIC} \cite{fan2023arctic}                  & 2.1M    (9)     & mocap           \\
    \midrule
    \rowA \multicolumn{3}{l}{\textcolor{darkgray}{w/ Estimated Contacts Annotation}} \\
    \textcolor{mygreen}{ContactPose} \cite{brahmbhatt2020contactpose} & 2.9M    (3)     & multi-kinect  \\
    \textcolor{mygreen}{GRAB} \cite{GRAB:2020}                        & -       (-)     & mocap         \\
    \textcolor{mygreen}{H2O} \cite{kwon2021h2o}                       & 571k    (5)     & multi-kinect    \\
    \midrule
    \rowA \multicolumn{3}{l}{\textcolor{darkgray}{w/ Ground-Truth Contacts Annotation}}  \\
    \textbf{\textcolor{mypink}{\DatasetName}}                     & \multirow{2}{*}{7M (50+)}    & \multirow{2}{*}{multi-auto}\\
    \textbf{\textcolor{mypink}{(Ours)}} \\
    \bottomrule
  \end{tabular}
  \caption{Dataset Comparison of existing Real World Datasets. The hands in previous datasets are represented by \textcolor{myblue}{skeleton} and \textcolor{mygreen}{MANO}. Different from other works, we use \textcolor{mypink}{Gaussian} to model the hand. The keyword ``single/multi-manual" denotes whether single or multiple views being used to annotate manually.}
  \label{tab:dataset_comparison}
  \vspace{-0.08in}
\end{table}
}

\section{Background}
\label{sec:background}
We briefly summarize recent advances in modeling radiance fields of static and dynamic scenes using 3D Gaussians~\cite{Kerbl20233DGS,luiten2023dynamic,wu20234dgaussians}.
Our method (see ~\Cref{sec:method}) extends the 3D Gaussians representation to articulated objects like the hand, and for grasp capture.

\parahead{Static 3D Gaussians}
Given multi-view images and a sparse point cloud of the scene, a set of 3D Gaussian primitives
can be defined across world space $x \in \mathrm{R}^{3\times1}$ as,
\begin{equation} 
   G(x) = e^{\frac{-1}{2} (x - \mu)^T \Sigma^{-1}(x - \mu)},\nonumber
\end{equation}

here each Gaussian primitive has 3D position ($\mu$), opacity, anisotropic covariance matrix ($\Sigma$), and spherical harmonic (SH) coefficients.
During the training of the radiance field, the properties of the initial 3D Gaussians are optimized together with a tile rasterizer~\cite{Kerbl20233DGS} with the objective of minimizing pixel loss.

\parahead{Dynamic 3D Gaussians}
The 3D Gaussians approach has recently been extended to dynamic scenes~\cite{Kerbl20233DGS,wu20234dgaussians}.
\cite{wu20234dgaussians} introduces a deformation field that tracks the Gaussian position across timesteps. 
Similarly, \cite{luiten2023dynamic} enable Gaussians to move and rotate over time while maintaining their color, opacity, and size.
While these methods can capture dynamic and deformable scenes, they do not provide a way to control dynamic motion, \eg,~using a skeleton.
Furthermore, in these methods, Gaussians are free to move within the scene without any restrictions, which isn't suitable for representing hands due to their kinematic structure.
An articulated 3D Gaussians representation would be advantageous for grasp capture since it would enable low-dimensional skeleton-based control of the hand.

\section{Method}
\label{sec:method}
\shortname aims to perform markerless capture of human hand grasps by accurately estimating the shape, appearance, and contacts between the hand and the object from multi-view RGB videos. 
We achieve this by combining \handmodel with an object model, both represented as 3D Gaussians, enabling us to compute contacts more efficiently than sampling-based implicit representations.
\Cref{fig:arg_mags} provides an overview of our method. 


\subsection{\handmodel}
\label{sec:handmodel}
Our template-free, articulated hand model \handmodel adopts 3D Gaussian splatting as the representation for accurate shape and appearance modeling of hands. 
Our model can be trained on sequences from any multi-view dataset to build an articulable hand model at any novel pose.

\parahead{Representation}
\handmodel (see \Cref{fig:arg_hand}) is composed of a skeleton with 21 bones and has 26 degrees of freedom (check supplementary for bone-specific DOFs).
We built a custom pose estimation pipeline that uses AlphaPose~\cite{fang2022alphapose} to estimate the 3D joint positions followed by an inverse kinematics fit (check supplementary).
Since bone lengths can vary among different individuals, we estimate these lengths from the dataset and adjust the skeleton accordingly.
The unique shape and appearance of a person's hand in a canonical pose are determined by the states of 3D Gaussians, \ie,~positions $\mu$, covariances $\Sigma$, opacities $\alpha$, and spherical harmonics coefficients $\phi$.
The covariance of each Gaussian in the canonical space is further defined as $\Sigma = R S S^T R$, 
where $R$ and $S$ denote the rotation and scaling of the Gaussians. 

\parahead{Optimization}
A unique \handmodel is optimized separately for each subject from a dense multi-view dataset containing approx 20 hand poses. 
To initialize Gaussian states in \handmodel, we set their means to be points on a normal distribution centered at the midpoint of each bone in a \emph{canonical} hand pose, with the distribution's standard deviation adjusted to match the bone's length (as shown in \Cref{fig:arg_hand} ).
We follow a similar protocol as \cite{Kerbl20233DGS} to initialize the covariances, opacity, and SH coefficients. 

To get the Gaussian positions in the posed space, forward kinematics and linear blend skinning is applied to the canonical Gaussians.
One way to obtain skinning weights is to assign MANO weights~\cite{MANO:SIGGRAPHASIA:2017} directly to the closest Gaussians.
However, this approach results in artifacts because Gaussians could move in unpredictable ways during training leading to mismatched skinning weights (visualized in ablation study)
To address this, we create a canonical grid inspired by Fast-SNARF~\cite{chen2023fast}. 
Skinning weights are then allocated to grid voxels using the nearest neighbor method, termed as grid weights.
Now to obtain the skinning weights for the queried Gaussians $W$ in the canonical space, trilinear interpolation of these grid weights is performed. 
We calculate the transformed Gaussian positions using a per-bone transformation matrix, denoted as $T_b$ and linear blend skinning:
    $T_g = W T_b$,
    $\mu_p = T_g \mu$,
where $\mu_p$ represents the location of Gaussians in the posed space, and $T_g$ represents the transformation matrix for each Gaussian.
To compute the covariance of the Gaussians in the posed space, it is transformed using a rotation matrix $R_g$, derived from $T_g$. This is expressed as
$\Sigma_p = R_g \Sigma R_g^T$.
Regarding the appearance, we optimize spherical harmonics coefficients for each Gaussian $\phi_g$ in the canonical space. 
To get the colors in the transformed or posed space, the view direction from posed space $\nu^g_p$ is first converted to the canonical space $\nu^g_c$ 
as 
$\nu^g_c = T_g^{-1} \nu^g_p$,
using $T_g$ for each Gaussian.
After this step, we use these transformed view directions $\mu^g_c$ to query the spherical harmonics coefficients in canonical space and get corresponding RGB colors for each posed Gaussian. 
To get the final image rendering, all Gaussian states currently in the posed space are used as inputs to a differentiable rasterizer~\cite{Kerbl20233DGS}, denoted as $\mathcal{R}$
\begin{equation}
\label{eqn:rendering}
\mathcal{I} = \mathcal{R}(\mu_p, \nu_c, \Sigma_p, \alpha, \phi),
\end{equation}
where $\mathcal{I}$ is the rendered image.
During optimization, the Gaussian states are optimized using to minimize pixel loss on the posed hand.
To optimize all Gaussian states, we impose a rendering loss $\mathcal{L}_1 = \lVert \hat{\mathcal{I}} - \mathcal{I} \rVert$
and structural similarity~\cite{wang2004image} loss $\mathcal{L}_{SSIM}$ 
between synthesized image $\mathcal{I}$ and ground truth image $\hat{\mathcal{I}}$ of the posed hand. 
To further improve the perceptual quality of the synthesized images, we add an additional perceptual loss $\mathcal{L}_{perc}$ \cite{johnson2016perceptual}. 

To avoid highly anisotropic Gaussians that could cause artifacts in the contact rendering, we incorporate an isotropic regularizer which ensures optimized Gaussians remain as isotropic as possible.
If $\min_s \in R^3 $ and $\max_s \in R^3$ are the minimum and maximum scale of the optimized Gaussians, then isotropic regularizer $\mathcal{L}_{iso}$ is defined as
\begin{equation}
\mathcal{L}_{iso} = (\frac{\min_s}{\max_s} - s)^2,
\end{equation} 
where $s$ is set to be 0.4.
Our final loss function is 
$L_h = \alpha \mathcal{L}_1 + \beta \mathcal{L}_{SSIM} + \gamma \mathcal{L}_{perc} + \delta \mathcal{L}_{iso}$.

\parahead{Inference}
Once the Gaussian states are optimized, we can drive \handmodel using a skeleton obtained from our pose estimation pipeline (check supplementary).
Given a novel pose during the inference, \handmodel outputs the transformed Gaussians as well as the rendered image from a particular view. 

\subsection{\shortname: Grasp Capture}
\label{sec:grasp_capture}
While \handmodel enables high-fidelity articulated hand modeling, it is not designed for capturing grasps and contacts.
To capture grasps, we need a representation of the object as well as a method to estimate contacts.

\parahead{Object Representation}
For accurate representation of objects, we build a non-articulated Gaussian representation following \Cref{sec:handmodel} with some improvements to maintain geometric consistency and accuracy.
To prevent floaters during optimization, we prune outlier Gaussians by projecting on image and culling if they lie outside the object mask.

\parahead{Grasp Capture}
To capture the grasp in a particular sequence, we first articulate \handmodel using the estimated hand pose.
We then construct the object model as described above.
Next, we combine both hand and object Gaussians.
More specifically, if $G_h$ and $G_o$ are the hand Gaussians and object Gaussians in the grasp scene, we simply concatenate the Gaussians $G_f = \{G_o , G_h\}$.
Because we use Gaussian Splatting, it allows such a concatenation operation naturally -- this would not be possible with implicit representations~\cite{corona2022lisa,li2022tava, mundra2023livehand}.
As the rasterization module only requires a set of Gaussians and their states, we can seamlessly merge hand and object Gaussians for every frame. 
The final grasp image is given by a rasterized composition of these Gaussians using \Cref{eqn:rendering}. 

\parahead{Contact Estimation} 
The contact map is calculated based on the proximity in 3D space between hand and object Gaussian positions. 
For each Gaussian on the hand, we find the closest Gaussian on the object. 
This pair is considered to be in contact if their distance is less than a certain threshold, and the same applies when assessing contact from the object's perspective. 
Specifically, if $G_h$ represents the Gaussians on the hand and $G_o$ those on the object in the posed space, then the 3D contact map between them is defined as:
\[
C = 
\begin{cases} 
d(G_h, G_o), & \text{if } d(G_h, G_o) < \tau \\
0, & \text{otherwise}
\end{cases},
\]
where $d$ represents the pairwise Euclidean distance between the Gaussian locations. A contact is considered to have occurred if this distance is less than $\tau$, which is the predefined threshold for contact.
We then use this method to estimate two kinds of contact maps on the hand and object: (1)~an \textbf{instantaneous contact map} that denotes contact at a specific timestep, and (2)~an \textbf{accumulated contact map} that denotes contact after the grasping has concluded.
To get the accumulated contact map $C_{acc}$ we simply add the previous frame's accumulated contact map to current frame. 
For rendering contact maps, we employ \Cref{eqn:rendering} 
using the contact distance as the color value of each Gaussian. 

\subsection{\DatasetName}
\label{sec:dataset}
For our grasp capture method to work well, a key requirement is a multi-view RGB dataset with
tens of camera views that help resolve self-occlusions.
Many prior datasets (see ~\Cref{sec:literature} and ~\Cref{tab:dataset_comparison}) contain multi-view images or video of hand grasps~\cite{simon2017hand,hampali2020honnotate,taheri2020grab}, but none have the large number of views needed to support neural field representations or are limited to hands only~\cite{moon2020interhand2}.
We therefore present \DatasetName, a large real-world multi-view RGB grasp dataset with over \~7M frames from 50+ cameras, providing full 360-degree coverage of grasp sequences comprising of 30+ diverse object scenes.

\parahead{Capture System}
Our customized data capture setup consists of 53 RGB cameras uniformly located inside a cubical capture volume with each cube face consisting of 9 cameras.
The sides of the cube are illuminated evenly using LED lights.
Each RGB camera records at 120~FPS with a resolution of $1280\times720$.
The cameras are software synchronized with a frame misalignment error of no more than 3~ms.
The multi-view system is calibrated for camera intrinsics and extrinsics using COLMAP~\cite{schoenberger2016sfm,schoenberger2016mvs} with fiducial markers on the walls.

\parahead{Capture Protocol}
Our capture protocol consists of four steps. First, we recorded multi-view videos of a subject's right hand as they performed a brief articulating movement.
Next, we capture only the object without the hand.
Then, without moving the object, we record multi-view videos of the subject's hand grasping the object.
We repeat this process 30+ times per subject with 2-5 grasps per object scene.
For evaluation sequences, we additionally capture a canonical pose at the end to record accumulated contacts seen in the transferred paint (see below).

\parahead{Ground Truth Contact}
A unique feature of our dataset is the capture of 15 evaluation sequences where the object has wet paint during the grasp~\cite{kamakura1980patterns}.
As a result, paint is transferred to the hand resulting in visual evidence of contact.
This contact mark is a physically accurate representation of the true (accumulated) contact between the hand and the object making it the true ground truth (even methods like \cite{brahmbhatt2019contactdb} suffer from heat dissipation).
We chose a bright green paint to enable automatic segmentation thereby creating a \textbf{gold standard} for contact evaluation. 

\parahead{Data Annotation}
\DatasetName also provides 2D and 3D hand joint locations along with hand and object segmentation masks.
We obtain the joint locations from AlphaPose~\cite{fang2022alphapose} followed by 3D triangulation and inverse kinematics~\cite{handtracker_iccv2013}.
We impose constraints to limit the degrees of freedom and joint angles for the rotation of the bones.
To achieve temporal smoothness for the sequence, we apply the 1\texteuro~Filter~\cite{Casiez20121F} on the estimated parameters. 
To segment the hand and object from the background, we use the Segment Anything Model (SAM)~\cite{kirillov2023segany} followed by fitting an Instant-NGP model~\cite{mueller2022instant} to extract a binary mask to ensure multi-view consistency. 
\section{Experiments and Results}
\label{sec:results}
In this section, we show qualitative and quantitative results from our method.
Our goal is to evaluate both the \handmodel and the \shortname grasp capture method, and compare with existing methods.

\subsection{Evaluating \handmodel}
%
\begin{figure}[h!]
  \centering
  \includegraphics[width=\linewidth]{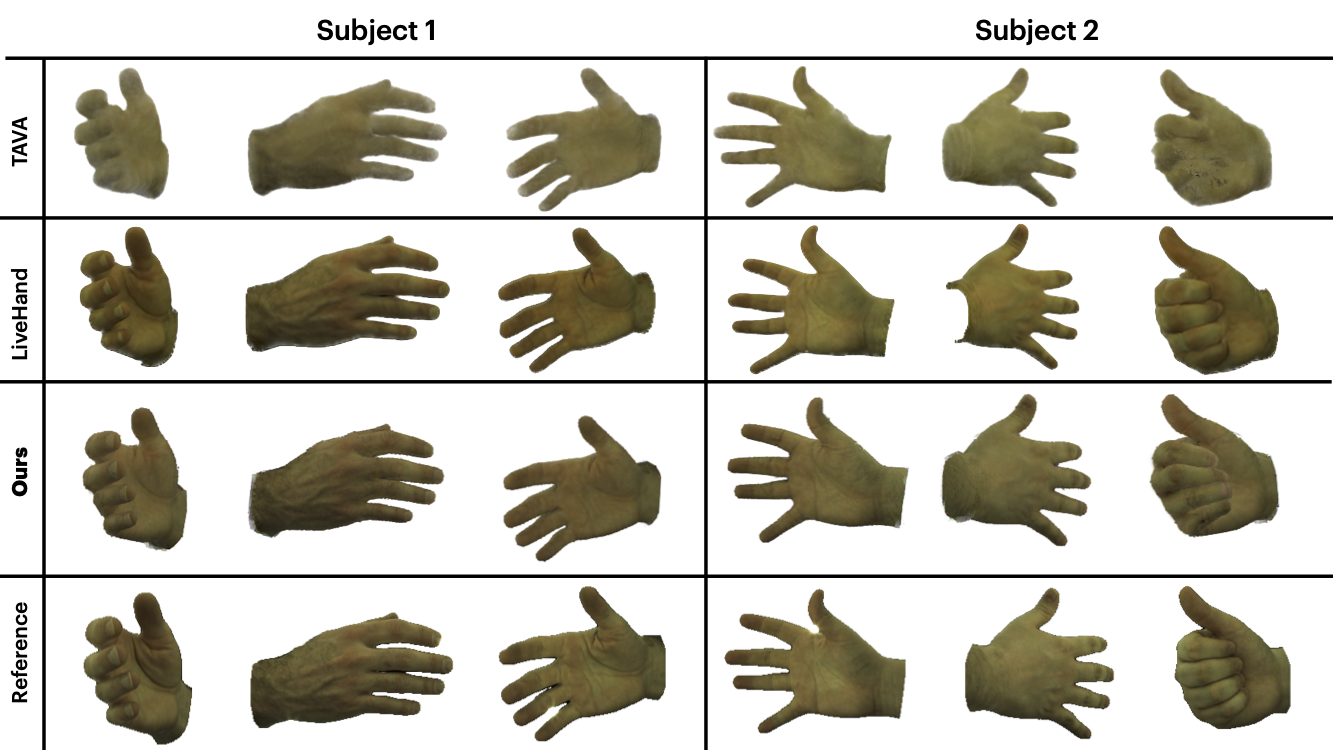}
  \caption{Qualitative comparison of \handmodel with LiveHand \cite{mundra2023livehand} and TAVA \cite{li2022tava}. It's noteworthy that our renderings closely resemble those of LiveHand and surpass TAVA in quality, even in the absence of any components designed to enhance photorealism.
  }
  \label{fig:hand_comparison}
\end{figure}
We first show results and experiments related to \handmodel only.
We quantitatively as well qualitatively assess the visual quality of our hand model with the current state-of-the-art method LiveHand~\cite{mundra2023livehand} and TAVA~\cite{li2022tava}.
\parahead{Metrics, Dataset \& Setup}
We assess the visual quality of our hand model using PSNR, SSIM, and LPIPS metrics (where higher scores indicate better performance) on the Interhand2.6M dataset, as shown in \Cref{tab:hand_quantitative}.
We used two subjects from Interhand2.6M (Capture0 and Capture1), focusing on the ``ROM07-RT-Finger-Occlusions" sequence from the test set. 
We allocate 75\% of the data for optimizing and use the remainder for evaluation.

\parahead{Quantitative Evaluation}
\handmodel is not specifically designed for photorealism since we leave out ambient occlusion and shadow mapping and focus only on geometric accuracy. 
As shown in \Cref{tab:hand_quantitative}, our results outperforms TAVA however LiveHand emerges as the best in terms of the evaluated metrics (PSNR/LPIPS), which significantly penalize the absence of ambient occlusion and shadows (also mentioned by \cite{li2022tava}). 
\textit{ We want to emphasize that our primary goal is not to surpass existing
hand models in terms of visual quality.
Instead, our focus is on accurate contact estimation.}
LiveHand and TAVA both learn implicit volumetric density field which makes calculating contact maps complicated \& expensive, whereas our Gaussians-based approach is more efficient. 
The comparison with LiveHand and TAVA is intended to demonstrate our comparable visual quality despite not being designed for it. 


\parahead{Qualitative Evaluation}
We conducted a qualitative comparison of our \handmodel with TAVA \cite{li2022tava} and LiveHand \cite{mundra2023livehand}, as shown in \Cref{fig:hand_comparison}. 
The quality of our renderings is superior to TAVA \cite{li2022tava} and is on par with that of LiveHand. 
In conclusion, despite not being tailored for photorealism, our method demonstrates substantial potential for application in photorealistic contexts.

\subsection{Evaluating Grasp Capture}
Next, we evaluate our \shortname method for grasp capture.
In this paper, we assume that direct contact between the hand and the object is the primary mode of grasping (we ignore indirect grasping through tools).
Therefore, the goal of grasp evaluation is to objectively measure the accuracy of contacts.
We compare three methods: (1)~ MANO~\cite{MANO:SIGGRAPHASIA:2017} fitting methods, (2)~HARP~\cite{Karunratanakul2022HARPPH}, and (3) our \shortname model.

\parahead{Metric, Dataset \& Setup}
In our experiments, we use the wet-paint transfer method ~\cite{kamakura1980patterns} to accurately collect ground truth accumulated contacts (see \Cref{sec:dataset}).
After grasp completion, users are instructed to return to a canonical post-grasp pose.
In this pose, the green paint residue in the grasping hand is automatically segmented and 2D contact maps are rendered from 10 different views (details in supplementary) using \cite{mueller2022instant}.
We then assess the quality of grasps estimated by different methods using the Intersection over Union (IoU) and F1-score metrics.
All experiments use 15 sequences of our wet-paint evaluation sequences.
We set the distance threshold $\tau = 0.004$ for contact estimation for all methods.
For a fair comparison, we subdivide the meshes of MANO and HARP from 778 to 49,000 vertices before estimating contact.
For estimating contact masks in all methods, we utilize the 'gray' color map \cite{Hunter:2007} on the distance map. 
The contact masks for \shortname are rendered using \cite{Kerbl20233DGS}, while for the other two frameworks, they are rendered using the emission shader in Blender.
It's noteworthy that \shortname \textbf{consistently outperforms} the others in the contact metric across all three subjects as shown in \Cref{tab:contacts_quantitative}.

\begin{table}[h!]
  \centering
  \begin{tabular}{cccc}
    \toprule
    {Method} & {Subject1}  & {Subject2}  & {Subject3}\\
    \midrule
    \rowA \multicolumn{4}{l}{\textcolor{darkgray}{mIoU $\uparrow$}} \\
    MANO & 0.161  & 0.135  & 0.208 \\
    HARP & 0.173  & 0.148 & 0.224  \\
    \textbf{Ours} &  \textbf{0.206 } & \textbf{0.152 } & \textbf{0.275 }  \\
    \midrule
    \rowA \multicolumn{4}{l}{\textcolor{darkgray}{F1 score $\uparrow$}} \\
    MANO & 0.270 & 0.228 & 0.338 \\
    HARP &0.28875 &  0.2474&  0.361 \\
    \textbf{Ours} &  \textbf{0.335} & \textbf{0.251} & \textbf{ 0.424}  \\
    \bottomrule 
  \end{tabular}
  \caption{Comparison of MANUS grasp capture approach with MANO and HARP on contact metric. Note that, we perform consistently better in both metrics.
  }
  \label{tab:contacts_quantitative}
\end{table}

\begin{figure*}[ht!]
  \centering
  \includegraphics[width=0.98\textwidth]{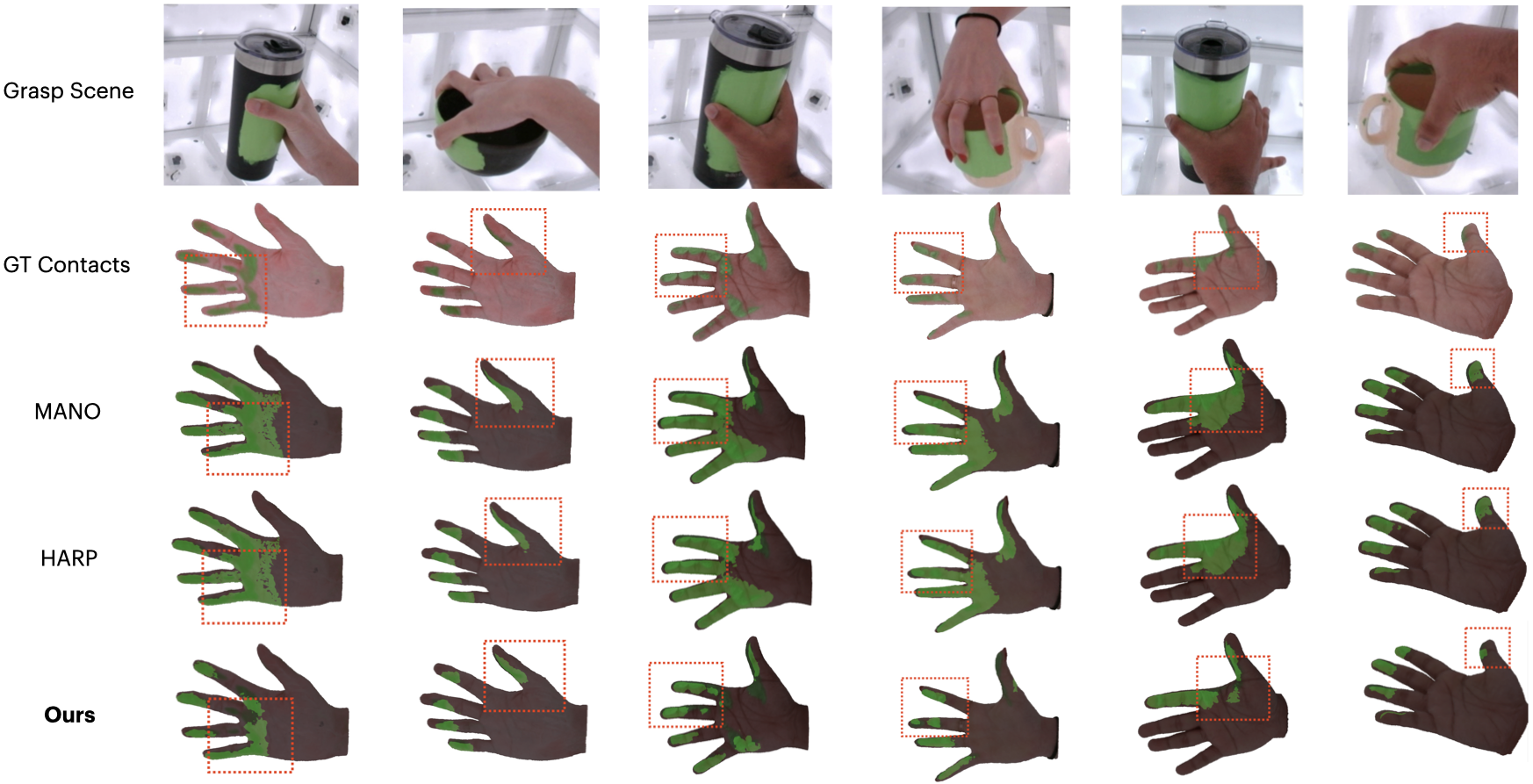}
  \caption{\textbf{Contact Comparisons}: We compare accumulated contacts of MANUS with that of MANO and HARP on ground truth contacts from MANUS Grasps dataset. It's visible that our contacts are far more accurate and closer to the actual ground truths.  
  }
  \label{fig:contact_comparison}
\end{figure*}

\parahead{Qualitative Evaluation}
We also present a qualitative comparison of our contact results against those obtained using MANO and HARP in \Cref{fig:contact_comparison}.
Our method shows a more accurate representation of the contact area, closely matching the actual contact masks, unlike the over-segmentation observed in MANO and HARP methods.
Although our method outperforms others, we note that there is still significant room for improvement on our dataset for future methods to address. 

\parahead{Discussion}
We also demonstrate the importance of dense camera views for accurate contact representation in \Cref{tab:contacts_ablation} which shows the diminishing of contact metric as the number of camera view decreases.
This finding is significant as it confirms our initial hypothesis that dense camera views are essential for accurate contact representation, helping to prevent self-occlusion scenarios.

\begin{figure*}[!ht]
  \centering
  \includegraphics[width=0.98\textwidth]{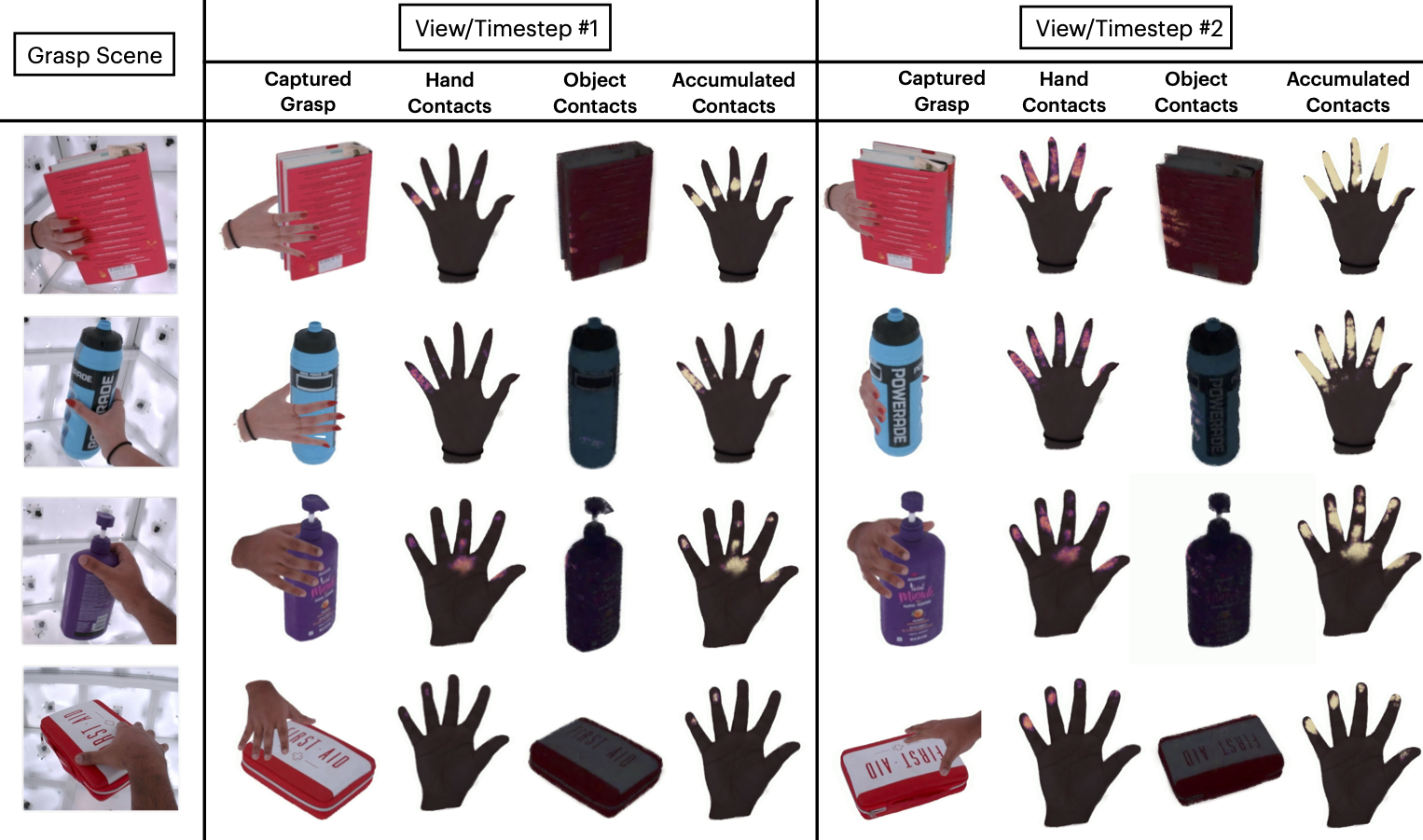}
  \caption{Here we show our contact estimation results on novel views for a variety of objects. We show both instantaneous and accumulated contacts for the hand in a canonical pose. Best viewed zoomed. }
  \label{fig:main_results}
\end{figure*}

\parahead{Results} Finally, we show qualitative results in \Cref{fig:main_results}, showcasing two different stages: one during the grasp process and another at the conclusion of the grasp.  
For a comprehensive 360-degree view of the grasp capture, an in-depth ablation study, and details on the implementation, please refer to our supplementary materials.
\section{Conclusion}
In this work, we proposed \shortname, 
which introduced a novel articulated 3D Gaussians representation, which successfully bridge the gap between the accurate modeling of contacts in hand-object interactions and the limitations of current data capturing techniques. 
We introduced \DatasetName, an extensive multi-view dataset captured from 50+ cameras, which offers an unprecedented level of detail and accuracy, covering a wide range of scenes, subjects, and frames. 
Overall, \shortname demonstrates remarkable potential in advancing the fields of robotics, mixed reality, and activity recognition, enabling the creation of more accurate robotic systems and enhanced virtual interactions.

\parahead{Limitations and Future Work}
While our focus in this paper was on accurate contact estimation, we recognize that the complexity of hand dynamics in everyday life extends far beyond what we have explored. 
Our current focus has been on modeling single-hand grasping with static objects, without delving into the pose-dependent non-linear deformation caused by skin stretching. 
Additionally, hand-object manipulation for longer time-frames is unaddressed in this work and can be a interesting direction for future works. 
We also observe that there is room for improvement in the metrics we propose for future work.
We also acknowledge the complexity and limited accessibility of our capture setup which motivates us to make dataset publicly available. \\

\begin{table}[t!]
  \centering
  \setlength{\tabcolsep}{3pt}
  \begin{tabular}{ccccc}
    \toprule
    \multicolumn{1}{c}{Method} & \multicolumn{1}{c}{PSNR $\uparrow$}  &  \multicolumn{1}{c}{SSIM $\uparrow$}  & \multicolumn{1}{c}{LPIPS $\downarrow$} & \multicolumn{1}{c}{Test time (s) $\downarrow$}\\
    \midrule
    TAVA   &     22.85 & 0.983 & 0.099 & 11.00 \\
    LiveHand   &     \textbf{31.16} & 0.9818 & \textbf{0.0278} & \textbf{0.022} \\
    \textbf{Ours}   &   26.32 & \textbf{0.9872} &  0.068 & 0.049     \\
    \bottomrule
  \end{tabular}
  \caption{Here, we show comparison of MANUS-Hand on InterHand2.6M \cite{moon2020interhand2} dataset with LiveHand \cite{mundra2023livehand} and \cite{li2022tava}. Note that our primary goal is to obtain accurate contacts, not visual quality.
  }
  \label{tab:hand_quantitative}
\end{table}


\begin{table}[t!]
  \centering
  \begin{tabular}{cccc}
    \toprule
    {Camera Views} & {Subject1}  & {Subject2}  & {Subject3}\\
    \midrule
    \rowA \multicolumn{4}{l}{\textcolor{darkgray}{mIoU $\uparrow$}} \\
    5 &  0.147 & 0.140 &  0.214\\
    10 & 0.164  & 0.145  & 0.256  \\
    20 &   0.176 &   0.142 &   0.261 \\
    \textbf{Ours (30+)} &  \textbf{0.206 } & \textbf{0.152 } & \textbf{0.275 }  \\
    \midrule
    \rowA \multicolumn{4}{l}{\textcolor{darkgray}{F1 score $\uparrow$}} \\
    5 &  0.244 & 0.235 &  0.343\\
    10 & 0.266 & 0.242 & 0.401 \\
    20 &  0.271 &  0.240 & 0.410\\
    \textbf{Ours (30+)} &  \textbf{0.335} & \textbf{0.251} & \textbf{0.424}   \\
    \bottomrule
  \end{tabular}
  \caption{ Here we show empirical findings demonstrating the decline in contact metric as the number of camera views decreases, leading to increased susceptibility to self-occlusions. 
  }
  \label{tab:contacts_ablation}
  \vspace{-0.05in}
\end{table}

\parahead{Acknowledgements}
This work was supported by NSF CAREER grant \#2143576, ONR DURIP grant N00014-23-1-2804, ONR grant N00014-22-1-259, a gift from Meta Reality Labs, and an AWS Cloud Credits award. We would like to thank George Konidaris, Stefanie Tellex, and Dingxi Zhang. 
Additionally, we thank Bank of Baroda for partially funding Chandradeep's travel expenses.

\clearpage

{
    \small
    \bibliographystyle{ieeenat_fullname}
    \bibliography{main}
}

\clearpage
\setcounter{page}{1}
\maketitlesupplementary

\begin{figure}[h]
  \centering
  \includegraphics[width=\linewidth]{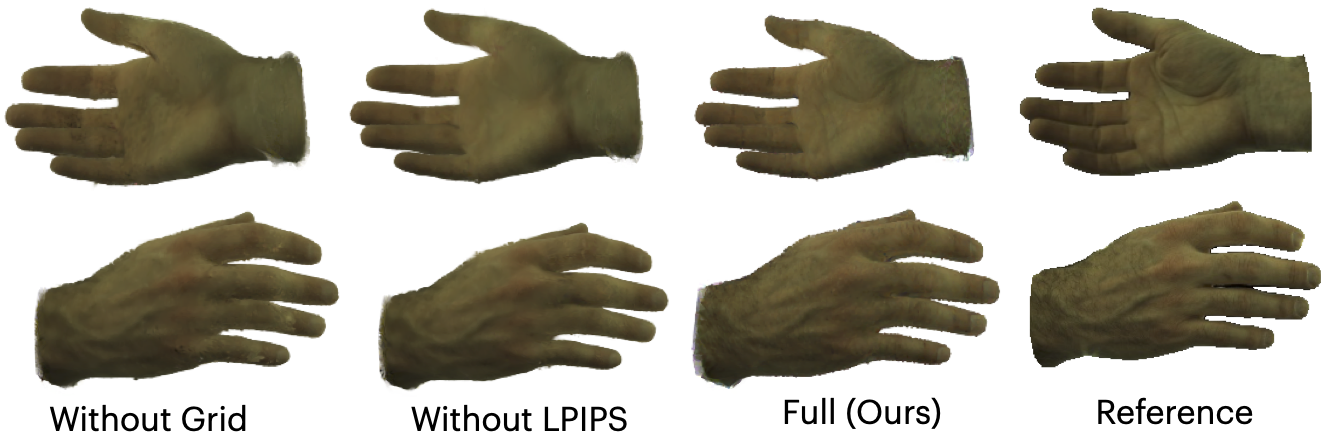}
  \caption{\textbf{Hand Ablation}: We perform ablation on the grid initialization of the skinning weights and the choice of LPIPS loss function. Clearly our approach is better in terms of visual appearance.}
  \label{fig:hand_ablation}
\end{figure}

\begin{figure}[h!]
  \centering
  \includegraphics[width=\linewidth]{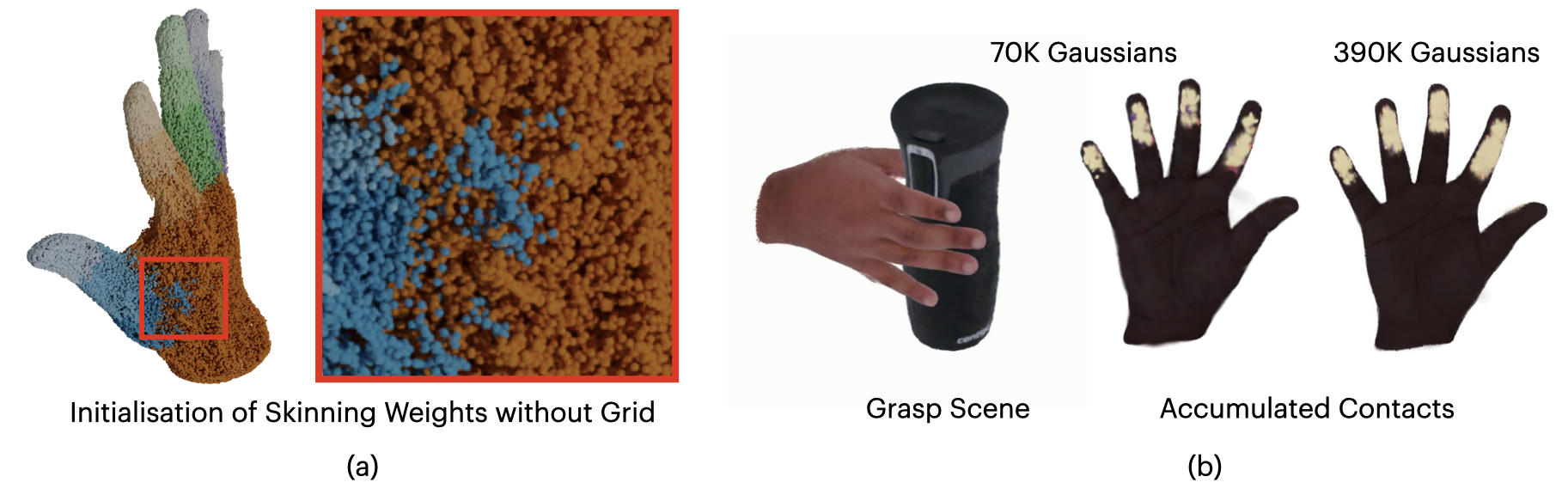}
  \caption{Here in \textbf{(a)} we show how initializing MANO weights without voxel grid allows the unstructured Gaussians to move erratically.  In \textbf{(b)}, we show the affect on accumulated 2D contact renderings with change in the number of Gaussians. }
  \label{fig:ablation_extra}
\end{figure}

\section{Ablation Study}
\subsection{\handmodel}
\parahead{Initialization of Skinning Weights}
We observe that the choice of method used to initialize skinning weights significantly influences the performance of our hand model. 
As demonstrated in \Cref{fig:ablation_extra} (a), initializing skinning weights directly onto Gaussians using a nearest neighbor approach, as opposed to grid initialization, leads Gaussians to move erratically and shift towards an unrelated bone.  
Consequently, this misalignment results in artifacts, where skinning weights are incorrectly allocated to the wrong bone, causing the position to be associated with the incorrect bone. 
The impact of this method of initialization is presented both quantitatively and qualitatively in \Cref{tab:hand_ablation} and \Cref{fig:hand_ablation}. 

\parahead{Ablation on LPIPS loss}
We observed that LPIPS loss improves the quality of renderings and maintain consistency across views. 
We also demonstrate that LPIPS loss function improves the overall visual quality of our hand model qualitatively at \Cref{fig:hand_ablation} and quantitatively at \Cref{tab:hand_ablation}. 

\begin{figure}[h!]
  \centering
  \includegraphics[width=\linewidth]{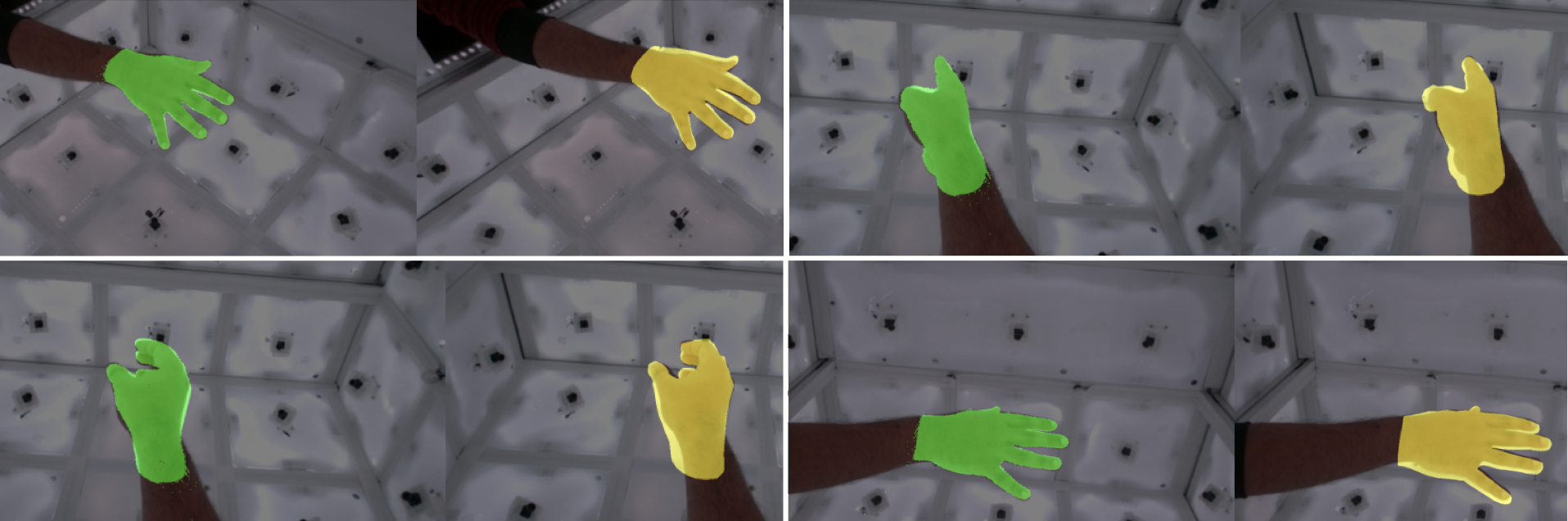}
  \caption{We display a comparison of the pixel misalignment between projected Gaussians and the MANO mesh against a reference image.}
  \label{fig:alignment}
\end{figure}

\parahead{Alignment with image pixels}
We now demonstrate the pixel-alignment results of MANUS-hand and MANO in \Cref{fig:alignment}. 
Due to inherent design and photo-metric losses, our hand representation is pixel-aligned to reference image, resulting in reduced alignment as compared to that of MANO. 

\parahead{Benchmarking MANUS Grasp scenes}
We also evaluate our MANUS Hand and Object method in \Cref{tab:dataset_benchmark} using the data included in the MANUS Grasp dataset.
The well-lit scenes and the absence of harsh shadows in our dataset lead to improved evaluation metrics when compared with those of the InterHand2.6M dataset. 

\subsection{\shortname Grasp Capture}
\parahead{Affect of the number of Gaussians in contact map rendering}
We show in \Cref{fig:ablation_extra}(b) that the quality of accumulated 2D contact maps deteriorates when the number of Gaussians is reduced. 
Therefore, in our experiments, we make sure to densely initialize Gaussians for both objects and hands.

\begin{table}[h]
  \centering
  \setlength{\tabcolsep}{3pt}
  \begin{tabular}{ccccc}
    \toprule
    \multicolumn{1}{c}{Method} & \multicolumn{1}{c}{PSNR $\uparrow$}  &  \multicolumn{1}{c}{SSIM $\uparrow$}  & \multicolumn{1}{c}{LPIPS $\downarrow$} & \multicolumn{1}{c}{Test time (s) $\downarrow$}\\
    \midrule
    w/o grid    & 26.108 &  0.987 &  0.0729 &  \textbf{0.0082} \\
    w/o lpips     & 25.92   &  0.986 &  0.074 & 0.043 \\
    \textbf{Ours}   & \textbf{26.328} &  \textbf{0.9872} & \textbf{0.0688} &  0.043  \\
    \bottomrule
  \end{tabular}
  \caption{Ablation on weight initialization approach and choice of LPIPS loss. Our design approach improve all visual quality metrics.  
  }
  \label{tab:hand_ablation}
\end{table}

\begin{table}[t!]
  \centering
  \begin{tabular}{ccccc}
    \toprule
    \multicolumn{1}{c}{Categories} & \multicolumn{1}{c}{PSNR $\uparrow$}  &  \multicolumn{1}{c}{SSIM $\uparrow$}  & \multicolumn{1}{c}{LPIPS $\downarrow$} \\
    \midrule
    Mugs    &   43.08  &   0.999  &  0.002 \\
    Bottles   &   38.17  &  0.997   & 0.008 \\
    Fruits   &   39.57  &  0.998   & 0.005 \\
    Utensils   &   38.25  &   0.994  & 0.009 \\
    Misc   &  38.79   &  0.995   & 0.008 \\
    Colored   &  42.38   &   0.999  & 0.004 \\
    Bags   &   38.44  &  0.994   & 0.011 \\
    Jars   &  40.66   &  0.999   & 0.005 \\
    Books   &  36.17   &   0.998  &  0.015   \\
    Tech   &  38.81   &   0.995   & 0.007 \\
    Hand1   &     28.34 &      0.995 & 0.031 \\
    Hand2   &     29.94 &      0.998 &  0.029 \\
    Hand3   &     29.71 &      0.997 &  0.027 \\
    \bottomrule
  \end{tabular}
  \caption{Here we benchmark MANUS-hand and object method on MANUS Grasp scenes.}
  \label{tab:dataset_benchmark}
\end{table}

\section{Implementation Details}
Our method was implemented in Python using the PyTorch Lightning~\cite{Falcon_PyTorch_Lightning_2019} framework. 
All experiments were conducted using a single Nvidia RTX3090 GPU with gradient accumulation for 4 iterations. 
The weights of the different loss function terms - $\alpha$, $\beta$, $\gamma$ and $\delta$ - were experimentally determined and set at values of 0.7, 0.1, 0.1, and 0.1, respectively. 
In all our experiments, we chose a grid size of 256x160x142 around the canonical hand skeleton for storing the skinning weights initialized from MANO~\cite{MANO:SIGGRAPHASIA:2017}.
\handmodel is initialized with 30K Gaussians per bone, amounting to 900K Gaussians in total.
After training, this number is pruned and filtered down to approximately 300K. 

\begin{figure*}[t!]
  \centering
  \includegraphics[width=\linewidth]{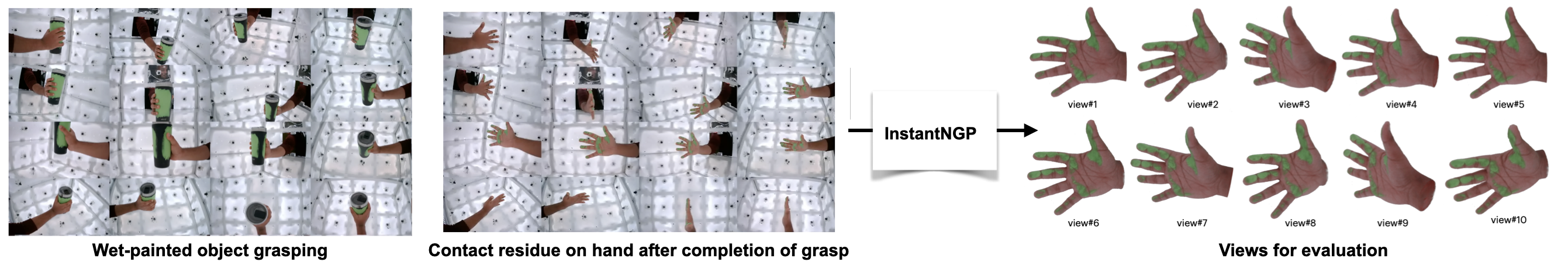}
  \caption{ Here, we show the approach we used to obtain the ground truth contacts for the evaluation sequences.
On the far right, we display all 10 views of one evaluation sequence for the quantitative assessment of grasp capture.}
  \label{fig:gt_contacts}
\end{figure*}

\section{\shortname Dataset Details}

\parahead{Bone length estimation}
We first use the \cite{fang2022alphapose} to acquire 2D keypoints for every frame and view. 
These keypoints are then triangulated into 3D keypoints using the \cite{easymocap}. 
With these triangulated keypoints, we determine the bone lengths for each subject. 
Specifically, we average the 3D keypoints across all grasp sequences and then adjust the length of the skeleton accordingly.

\begin{figure}[h]
\centering
\includegraphics[height=1.5in]{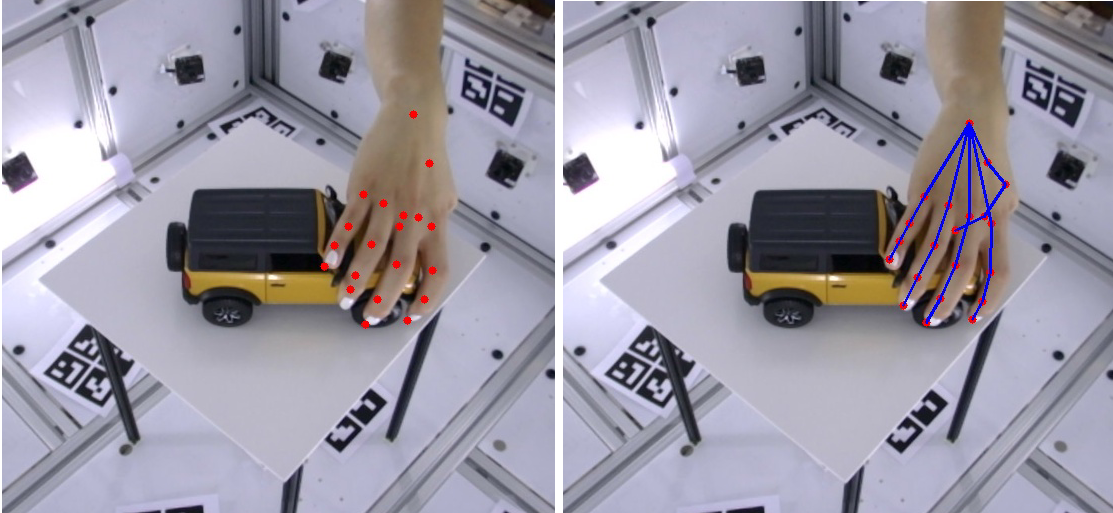}
\caption{The left figure shows the backprojected 3D keypoints predicted by AlphaPose \cite{fang2022alphapose}. The right figure shows the fitted hand skeleton using inverse kinematics.}
\label{fig:ik_figure}
\end{figure}

\begin{figure}[h]
  \centering
  \includegraphics[width=\linewidth]{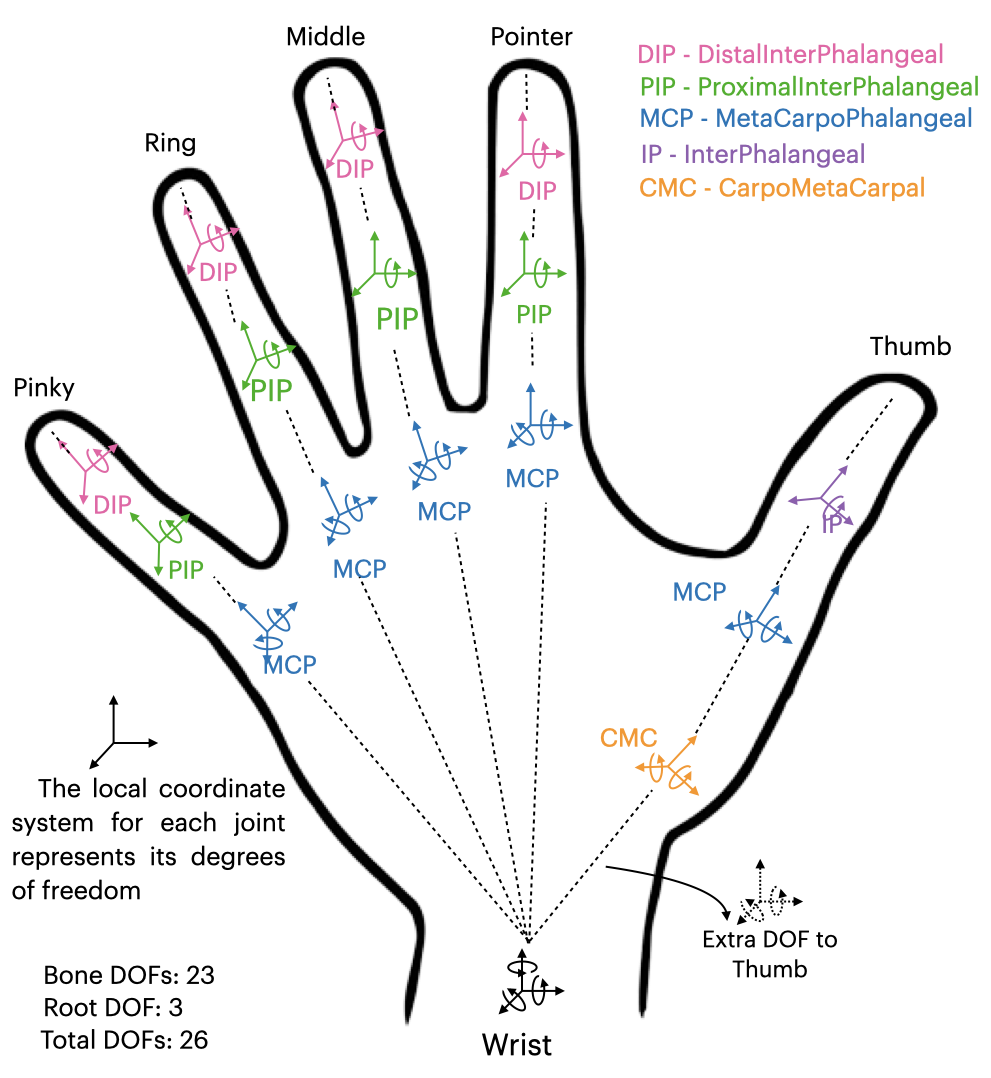}
  \caption{Figure showing the degrees of freedom of rotation for each of the joint.}
  \label{fig:dof_figure}
\end{figure}

\parahead{Inverse Kinematics}
To obtain the joint angles of the hand and its global orientation we use an optimization-based approach inspired by \cite{handtracker_iccv2013}. Specifically, we treat the joint angles, global rotation and global translation as optimization parameters $\Theta$. We then perform a forward kinematics ($Fk(\Theta)$) pass which takes the joint angles as input and outputs 3D joint locations. As the forward pass is differentiable, we apply gradient descent to obtain the optimal parameters that explain the given 3D joint positions. We minimize the L2 loss between predicted and target keypoints:
\begin{equation} \label{eqn:keyp_loss}
    \mathcal{L}_{kyp} = || Fk(\Theta) - x ||^2
\end{equation}
where $x$ are the 3D joint locations predicted by AlphaPose \cite{fang2022alphapose}.
We also impose anatomical constraints (See \Cref{fig:dof_figure}) and joint angle limits by applying a hinge loss as limit loss $\mathcal{L}_{lim}$ as follows:
\begin{equation}
    \mathcal{L}_{lim} = \sum_{i=1}^{|\Theta|} ((\max(0, ||\Theta^i - l_h^i||^2) + \max(0, || l_l^i - \Theta^i ||^2))
\end{equation}
where $l_l$ and $l_h$ are the lower and upper limits on joint angles, respectively.
The final loss function is given by:
\begin{equation}
    \mathcal{L} = \mathcal{L}_{kyp} + \lambda \mathcal{L}_{lim}
\end{equation}
We use Adam \cite{Kingma2014AdamAM} as our choice of optimizer with a learning of 0.001 and set the value of $\lambda$ to be 1. 
We also initialize the current frame based upon previous frame, this helps in faster convergence and helps in maintaining temporal consistency.
Once we get the joint angles, we apply one euro filter \cite{Casiez20121F} to the joint angles to smoothen any high-frequency jitter in the sequence. 
We show illustration of this process in \Cref{fig:ik_figure}. 

\parahead{Segmentation}
For every segmentation task, we employ a combined approach utilizing InstantNGP \cite{mueller2022instant} and SAM \cite{kirillov2023segany}. 
Initially, the scene is segmented using the text-based SAM technique. 
Following this, we obtain a segmentation mask that maintains consistency across multiple views using InstantNGP. 
If the segmentation masks are found to be inadequate due to inaccurate predictions from the text-based SAM, the process is repeated until satisfactory results are achieved.

\parahead{Ground Truth Contacts}
In ~\Cref{fig:gt_contacts}, we illustrate the methodology used to gather ground truth contact data for our evaluation sequences. Initially, the object is coated with a layer of bright, wet paint.
Following this, the object is grasped, resulting in the transfer of paint residue to the hand.
After the grasp is finalized, we document the pattern of contact residue left on the hand.
To obtain the required viewpoints, we train ~\cite{mueller2022instant} in the multi-view images and then select 10 distinct views for evaluation.
We repeat this process for 15 different evaluation sequences for each subject.

\begin{figure}[h!]
  \centering
  \includegraphics[width=\linewidth]{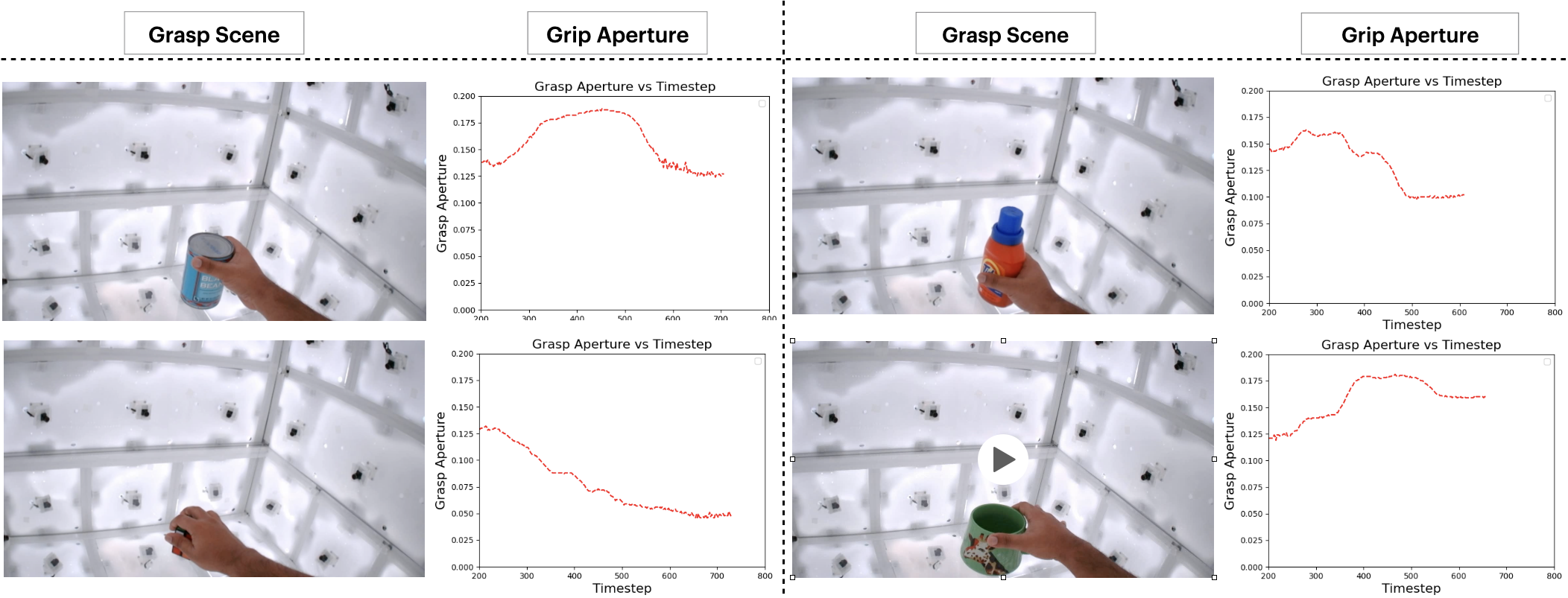}
  \caption{Variation of grip aperture with change in timestep while grasping. }
  \label{fig:grip_aperture}
\end{figure}

\parahead{Grip Aperture}
The grip aperture \cite{Castiello2005TheNO} refers to the distance between the thumb and fingers when grasping or holding an object. 
It's an important concept in fields like ergonomics, rehabilitation, and robotics. 
Here in \Cref{fig:grip_aperture}, we plot the change of grip aperture with change in timestep for our dataset. 
 
\section{MANO and HARP evaluation}
\parahead{Pose and Shape Estimation}
We begin by estimating the shape and scale parameters of the MANO model for each subject. 
First, we obtain the mesh for every time-step by training \cite{mueller2022instant} on multi-view images. 
Next, we refine the mesh through the use of MeshLab and Blender software to achieve a cleaned version. 
We employ an optimization framework akin to that used in \cite{hasson19_obman}, focusing on optimizing all MANO parameters, including angle, translation, shape, and scale for the first timestep.
This optimization incorporates both keypoint loss \eqref{eqn:keyp_loss} and point-to-surface loss \cite{ravi2020pytorch3d} with the clean mesh.
For subsequent sequences , we keep the shape and scale parameters unchanged, focusing solely on optimizing angles and translations through keypoint loss. 
To enhance the speed of convergence, we use the optimized parameters from the previous step as the starting point for new parameters.

To get better geometry than MANO we extend HARP ~\cite{Karunratanakul2022HARPPH} from monocular video setup to multi-view video setup. 
We start with already optimized MANO model (as mentioned above) and then optimize for the local displacement of the hand shape. 
We leverage the differentiable rasterizer, to optimize the HARP model based on the losses mentioned in ~\cite{Karunratanakul2022HARPPH}. 

\parahead{Evaluation Setup}
Please note that, we can't directly render contact maps for MANO and HARP in the same way as MANUS, which employs a Gaussian-based differentiable rasterizer.
To obtain contact maps for MANO and HARP, we initially allocate contact values to each vertex, followed by utilizing Blender's emission renderer to render the contact mask. 
For fair comparison, we increase the resolution of MANO and HARP vertices from 778 to 49,000.

\end{document}